\documentclass[11pt]{article}

\usepackage[preprint]{acl}

\usepackage{times}
\usepackage{latexsym}

\usepackage[T1]{fontenc}

\usepackage[utf8]{inputenc}

\usepackage{microtype}

\usepackage{inconsolata}

\usepackage{graphicx}

\usepackage{float}
\usepackage{amsmath}
\usepackage{amssymb}
\usepackage{booktabs}
\usepackage{placeins}
\usepackage{array}
\usepackage{listings}
\usepackage{fvextra}
\usepackage{multirow}

\usepackage{titlesec}
\titlespacing*{\paragraph}{0pt}{1.0ex plus 0.1ex minus 0.2ex}{1em}
\title{What Do Multilingual Self-Supervised Speech Encoders Know About Sentence Meaning?}

\author{Kevin Krahn \and Eric Fosler-Lussier \\
        Dept.\ of Computer Science and Engineering \\
        The Ohio State University \\
        \texttt{\{krahn.6, fosler-lussier.1\}}@osu.edu }

\begin{document}
\maketitle
\begin{abstract}
Self-supervised speech models (S3Ms) have been scaled up to thousands of languages and millions of hours of audio, yet little is known about how these models structure semantic knowledge across languages.
In this work, we probe sentence-level semantics in massively multilingual S3Ms across 76 languages, using linear maps from frozen speech representations into a frozen text embedding space, evaluated by cross-modal retrieval. Sentence meaning is linearly accessible but highly uneven across languages and resource tiers.
Cross-lingual probes can match language-specific probes on languages never seen in probe training, but per-model monolingual decodability does not predict transfer.
Within the OmniASR wav2vec family, scaling from 300M--7B disproportionately benefits high-resource languages, and ASR tuning sharply improves semantic decodability while making representations more language-specific.
However, semantic content is entangled with acoustics, inflating apparent performance on single-speaker corpora.
\end{abstract}

\section{Introduction}
Self-supervised speech models (S3Ms), trained on unlabeled audio, have become a standard representation layer for speech processing, and recent efforts have scaled them to hundreds or even thousands of languages and over a million hours of audio \citep{DBLP:conf/interspeech/ConneauBCMA21, DBLP:conf/interspeech/BabuWTLXGSPSPBC22, DBLP:journals/jmlr/PratapTSTBKENVF24, chen-etal-2024-towards-robust, omnilingualasrteam2025omnilingualasropensourcemultilingual}. Yet our understanding of what these models encode has not kept pace with this increase in language coverage.

Prior probing studies have established that S3M representations are dominated by acoustic and phonetic structure \citep{DBLP:conf/asru/PasadCL21, pasad-etal-2024-self, DBLP:conf/interspeech/ChoiPNFL024, DBLP:conf/interspeech/ErsoyMCADD25}, but also encode word \citep{DBLP:conf/asru/PasadCL21} and sentence \citep{pasad-etal-2024-self, DBLP:conf/icassp/ChoM0BA24, DBLP:journals/access/AshiharaDIK24} meaning. However, this work is limited to a small number of languages and usually monolingual models. Multilingual probing has focused on acoustic and phonetic unit discovery \citep{ortiz-tandazo-etal-2026-maubert, poli2026discophonbenchmarkingunsuperviseddiscovery} and downstream tasks such as automatic speech recognition (ASR) and language identification \citep{li-etal-2023-quantitative, DBLP:conf/interspeech/ShiBCHHCC0ML023, DBLP:conf/interspeech/ShiWCBKTCJLL024}. Multilingual S3Ms remain underexplored at the semantic level; it is unclear whether semantic content is present beyond high-resource languages or how it is structured across languages.

\begin{figure}[t!]
    \centering
    \includegraphics[width=1.0\columnwidth]{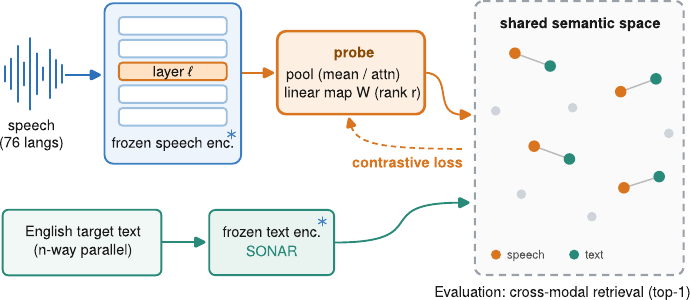}
    \caption{Overview of the probing approach. A linear probe maps multilingual speech representations into a text embedding space, where retrieval measures semantic decodability and zero-shot cross-lingual transfer.}
    \label{fig:method-overview}
\end{figure}

In this work, we present a sentence-level semantic probing study of massively multilingual S3Ms, covering 9 self-supervised models (plus several ASR-tuned variants) spanning training objective, scale (300M$\rightarrow$7B), and language coverage, across 76 languages. We train lightweight probes (Fig.~\ref{fig:method-overview}): learnable linear maps from frozen speech representations into a frozen  sentence embedding space with translation equivalence, keeping the maps linear to measure content already present rather than capacity added by the probe. We evaluate by cross-modal retrieval on $n$-way parallel corpora, where success requires recovering sentence meaning.

This setup separates two properties: \emph{semantic decodability} is how much
translation-equivalent meaning a linear map from speech to text space can recover. \emph{Cross-lingual alignment} is
whether that meaning is organized in a shared, language-agnostic fashion, allowing for cross-lingual transfer, which we test by varying probe training and test languages.

In summary, our findings are:
(i) Sentence meaning is linearly accessible in multilingual S3Ms but is highly uneven across languages and resource tiers.
(ii) Genuine cross-lingual semantic transfer exists; OmniASR cross-lingual probes can
match language-specific probes (zero-shot).
(iii) Monolingual decodability does not predict cross-lingual transfer; XEUS, with rich within-language semantics, has weak cross-lingual transfer.
(iv) Retrieval scores across OmniASR models ranging from 300M$\rightarrow$7B show that scaling up mostly benefits high-resource languages.
(v) ASR supervision strengthens semantic decodability, but in language-specific directions.
(vi) Semantics remain entangled with acoustics, inflating apparent decodability on single-speaker corpora.

\section{Related Work}

\subsection{Analysis of Self-Supervised Speech Representations}

Learning linguistic information from raw audio has long been a goal of the speech community, motivating
efforts such as the Zero Resource Speech Challenge
\citep{DBLP:journals/jstsp/DunbarHD22}. Such a capability would let
models serve communities whose languages are largely unwritten.
The foundational work wav2vec~2.0 \citep{DBLP:conf/nips/BaevskiZMA20}
popularized self-supervised learning from raw audio, but these models struggle with
semantic tasks that text models solve easily. \citet{DBLP:journals/jstsp/MohamedLBHEIKLL22}
note that even the original BERT corpus is 3B+ words (small by current
standards) whereas a comparable speech corpus would require on the
order of 400K hours of audio; this data gap is one reason text
models are often used to inject semantics into speech
models \citep{xu-etal-2023-introducing,
DBLP:conf/icassp/MaekakuSCF024}.

Furthermore, S3M representations tend to be more
phonetic than semantic \citep{DBLP:conf/interspeech/ErsoyMCADD25}, placing
phonetically similar words closer than synonyms under
cosine distance \citep{DBLP:conf/interspeech/ChoiPNFL024}.
Nonetheless, some semantic structure is
present: \citet{DBLP:conf/asru/PasadCL21} find that the middle, most
contextualized layers of wav2vec~2.0 best encode word meaning, and
\citet{pasad-etal-2024-self} show that S3Ms
carry sentence-level semantic content beyond word
identity, correlating with human judgments on a spoken semantic textual
similarity task. \citet{DBLP:conf/icassp/ChoM0BA24} similarly find, via
a spoken sentence ABX task, that wav2vec~2.0, HuBERT, and WavLM
distinguish semantically related English sentences from distractors.
But overall, these models lag substantially behind text models on spoken language understanding tasks
\citep{DBLP:journals/access/AshiharaDIK24}.

This evidence, however, is drawn mostly from monolingual models, while multilingual analyses such as ML-SUPERB \citep{DBLP:conf/interspeech/ShiBCHHCC0ML023, DBLP:conf/interspeech/ShiWCBKTCJLL024} evaluate across many languages but target downstream tasks (e.g., ASR and language identification), without exploring cross-lingual semantic structure.

\subsection{Supervised Speech-Text Alignment}

Past work has fine-tuned S3Ms
to align their representations with a pre-trained text embedding space,
turning them into semantic encoders. SAMU-XLS-R \citep{DBLP:journals/jstsp/KhuranaLG22} and SENSE models \citep{DBLP:conf/asru/MdhaffarECNE25}
distill a multilingual text encoder into a speech model at the utterance level through student--teacher alignment. SONAR
\citep{duquenne2023sonarsentencelevelmultimodallanguageagnostic} and Omnilingual SONAR
\citep{omnilingualsonarteam2026omnilingualsonarcrosslingualcrossmodal} similarly train a
shared language- and modality-agnostic sentence embedding space by aligning a
speech encoder to a frozen text encoder.

These models establish that a shared cross-lingual semantic space is
reachable from speech with explicit supervision, but they say
little about what is already latent in the frozen self-supervised
model. By probing frozen representations directly, we characterize the
semantic structure already present in S3Ms that text supervision can exploit.

\section{Probing Framework}

\subsection{Models}
\label{sec:models}

We evaluate 9 self-supervised speech models (Table~\ref{tab:models}) of different
architectures and training objectives:
the wav2vec~2.0 contrastive family \citep{DBLP:conf/nips/BaevskiZMA20},
scaled multilingually by MMS
\citep{DBLP:journals/jmlr/PratapTSTBKENVF24} and the Omnilingual (OmniASR) family
\citep{omnilingualasrteam2025omnilingualasropensourcemultilingual}; HuBERT-style masked
prediction \citep{DBLP:journals/taslp/HsuBTLSM21} in mHuBERT-147
\citep{DBLP:conf/interspeech/BoitoILBC24}; WavLM-style denoising
\citep{DBLP:journals/jstsp/ChenWCWLCLKYXWZ22} in XEUS
\citep{chen-etal-2024-towards-robust}; and the combined
contrastive--masked prediction objective of w2v-BERT-2.0
\citep{DBLP:conf/asru/ChungZHCQPW21,
communication2023seamlessm4tmassivelymultilingual}.
The OmniASR-W2V (wav2vec~2.0) encoders cover the 300M--7B
parameter range, isolating the effect of model size, and have ASR
CTC and LLM-style decoder variants,
letting us measure how transcription supervision affects semantic structure.
Finally, we include WavLM Base+ \citep{DBLP:journals/jstsp/ChenWCWLCLKYXWZ22}
as a strong English monolingual baseline.

\begin{table}[t!]
\centering
\small
\setlength{\tabcolsep}{2.7pt}
\begin{tabular}{>{\scriptsize}l r r r l r r}
\toprule
{\small \textbf{Model}} & \textbf{Params} & \textbf{$L$} & \textbf{$d$} & \textbf{Obj.} & \textbf{Hours} & \textbf{Langs} \\
\midrule
WavLM Base+   & 95M  & 12  & 768  & MP+D  & 94K   & 1     \\
mHuBERT-147   & 95M  & 12  & 768  & MP    & 90K   & 147   \\
XEUS          & 577M & 19  & 1024 & MP+D  & 1.1M  & 4057  \\
w2v-BERT-2.0  & 580M & 24  & 1024 & C+MP  & 4.5M  & 143+  \\
MMS-1B        & 965M & 48  & 1280 & C     & 491K  & 1406  \\
OmniASR-300M  & 317M & 24  & 1024 & C     & 4.3M  & 1600+ \\
OmniASR-1B    & 965M & 48  & 1280 & C     & 4.3M  & 1600+ \\
OmniASR-3B    & 3.0B & 60  & 2048 & C     & 4.3M  & 1600+ \\
OmniASR-7B    & 6.5B & 128 & 2048 & C     & 4.3M  & 1600+ \\
\bottomrule
\end{tabular}
\caption{Speech encoders probed in this work. $L$ = number of
transformer layers, $d$ = hidden dimension.
Objectives: C = contrastive (wav2vec~2.0), MP = masked prediction
(HuBERT-style), D = denoising (WavLM-style). Hours and languages refer
to pre-training data.
}
\label{tab:models}
\end{table}

\subsection{Datasets}
\label{sec:datasets}


We evaluate on two $n$-way parallel corpora differing in domain and speaker composition.


\paragraph{FLEURS.}
FLEURS \citep{DBLP:conf/slt/ConneauMKZADRRB22} provides multi-speaker
read-speech recordings of the FLORES machine translation benchmark
\citep{goyal-etal-2022-flores}, whose English source sentences are
drawn from Wikipedia. The material covers
science, geography, and current events, and is dense in named entities.
The dataset contains 1509, 150, and 350 sentences for train, validation, and test, respectively, averaging 2.3 recordings per sentence with different speakers.
We use a subset of 40 languages covering multiple language families, averaging
2591, 334, and 726 utterances per language for training, validation, and testing.
The sentences and speakers in each split are disjoint, so FLEURS scores measure cross-speaker generalization.

\paragraph{Bible.}
For the long tail of low-resource languages, scripture and other religious recordings are among the
only available speech data, constituting a substantial fraction of the
pre-training data in massively multilingual S3Ms,
extending to 1{,}300 languages in MMS \citep{DBLP:journals/jmlr/PratapTSTBKENVF24} and 4{,}000+ languages in XEUS \citep{chen-etal-2024-towards-robust}.
Bible data is therefore more in-domain than FLEURS for many low-resource languages.

We combine verse-segmented New Testament (NT) recordings from OpenBibleTTS
\citep{guzman2026openbiblettslargescalespeechresources}, Snow Mountain
\citep{raju2023snowmountaindatasetaudio}, and the English WEB version,
covering 48 languages with a single speaker per language, 12 of which are shared with the FLEURS subset.\footnote{See Appendix~\ref{appendix:resource_tiers} for per-dataset language breakdown.} This NT corpus primarily covers the mid- to low-resource range, including severely resource-constrained Himalayan and Omotic languages.
We use 4{,}000, 1{,}000, and 1{,}400 verses for training, validation, and testing;
six NT books are reserved for testing,\footnote{Acts, Gal, Phil, Col, 1 Thess, 1 John} so train and test are disjoint at the book level. Each language uses the same verse set, but missing or invalid verses ($<$1s or near-silent) are replaced with other verses (affecting $3.4\%$ total), keeping the corpus close to $n$-way parallel.

\subsection{Probe Structure}

We train probes that map the frame-level representations of a frozen S3M to a single vector in the embedding space of a frozen multilingual text encoder.
We embed the English text for each utterance to yield a single semantic target space shared by all probes.
We choose SONAR \citep{duquenne2023sonarsentencelevelmultimodallanguageagnostic} as our text encoder, because it is trained to map semantically similar inputs across languages to nearby vectors in a shared space, mitigating English bias compared to a monolingual text encoder.

Each probe maps a speech representation into the text encoder space; retrieval success then requires recovering sentence meaning, since the target contains no acoustic, speaker, or lexical overlap with the input. English is the exception; an English probe has a possible direct lexical mapping shortcut without requiring a translation step. Hence, we treat English as an upper-bound reference rather than evidence of semantic abstraction.

Given an utterance, the speech encoder produces frame-level hidden states $\mathbf{h}^{(\ell)}_1,\ldots,\mathbf{h}^{(\ell)}_T \in \mathbb{R}^{d}$ at layer $\ell$. We train probes on
every layer, including the initial convolutional output (layer~0). The text encoder maps the corresponding transcript to embedding $\mathbf{y} \in \mathbb{R}^{e}$. The probe consists of a pooling step, reducing $T$ frames to a single vector, followed by a learned linear projection into the text space.
Pooling computes a weighted sum of the $T$ frames,
\begin{equation}
\bar{\mathbf{h}}^{(\ell)} = \textstyle\sum_{t=1}^{T}\alpha_t\,\mathbf{h}^{(\ell)}_t
\end{equation}
where $\alpha_t$ is the weight for frame $t$. Our primary
results use mean pooling ($\alpha_t = 1/T$), but to measure whether frames carry
unequal semantic signal, we also compare with lightweight attention pooling which computes the
weights $\alpha_t$ with a single learnable query vector:
\begin{equation}
\alpha_t = \operatorname*{softmax}_{t}\!\big(\mathbf{w}_a^{\top}
\operatorname{LayerNorm}(\mathbf{h}^{(\ell)}_t)\big),
\end{equation}
where $\operatorname{softmax}$ normalizes the scores over the $T$
frames. This adds only $3d$ parameters ($\mathbf{w}_a \in
\mathbb{R}^d$ and the LayerNorm gain and bias), and we initialize
$\mathbf{w}_a$ to zero so that training starts from exact mean pooling,
deviating from uniform weighting where it provides benefit.
The pooled vector is mapped into the text space by a linear layer:
\begin{equation}
\mathbf{z} = \mathbf{W} \operatorname{LayerNorm}\!\big(\bar{\mathbf{h}}^{(\ell)}\big) + \mathbf{b}
\end{equation}
where $\mathbf{W} \in \mathbb{R}^{e \times d}$ and $\mathbf{b} \in \mathbb{R}^{e}$.
For a model with $d{=}1024$ (varies by model) and $e{=}1024$ for SONAR, the probe has roughly 1.05M trainable parameters, the vast majority in $\mathbf{W}$. Because a map of this
size potentially has the capacity to memorize acoustic-to-word associations instead of decoding semantic structure, we also train \emph{low-rank} probes in which $\mathbf{W}$ is factored as
$\mathbf{W} = \mathbf{A}\mathbf{B}$ with
$\mathbf{A} \in \mathbb{R}^{e \times r}$,
$\mathbf{B} \in \mathbb{R}^{r \times d}$, for $r < \min(d, e)$.

\subsection{Probe Training}
\label{sec:probe_training}
The probe is trained to minimize speech-to-text contrastive loss, matching the evaluation direction. Let
$\mathbf{y}_1,\ldots,\mathbf{y}_N$ be the embeddings of all $N$
training transcripts, and let
$Q_{ij} = s \cdot \cos(\mathbf{z}_i, \mathbf{y}_j)$ be the scaled
cosine similarity between speech embedding $\mathbf{z}_i$ and
transcript embedding $\mathbf{y}_j$, with fixed $s{=}20$.
For each utterance in a mini-batch we contrast its embedding
against the pool of training transcripts:
\begin{equation}
\mathcal{L} = -\frac{1}{B}\sum_{i=1}^{B}
\log \frac{\sum_{j \in P_i} \exp(Q_{ij})}
          {\sum_{j=1}^{N} \exp(Q_{ij})},
\end{equation}
where $P_i$ is the set of transcripts matching utterance $i$.
Parallel utterances in different languages share a target (for multilingual probes), and $P_i$ treats all instances of the matching transcript as positives.

We use the AdamW optimizer, a learning rate of $3 \times 10^{-3}$ ($1 \times 10^{-3}$ for attention pooling), and a batch size of 512. Weight decay has little effect on results, so we leave it at 0. Each probe is evaluated on the validation set every 15 steps, with early-stopping patience of 10 evaluations. Our final probe sweep used \textasciitilde 150 NVIDIA H200 GPU-hours.

\subsection{Probe Evaluation}
We embed all test-set English sentences and, for each test
utterance, retrieve the nearest text embedding by cosine similarity. We
report top-1 accuracy per-language and averaged over languages.\footnote{See Appendix~\ref{appendix:probe_seed} for variability tests over multiple seeds.}

\paragraph{Probe Regimes.}
We experiment with three main probe regimes that differ only in which languages supply
the probe's training data.
\textbf{Monolingual}: one probe per language, trained and evaluated on
that language, measuring within-language linear decodability.
\textbf{Multilingual}: a single shared probe trained jointly on all
languages, measuring whether one map can serve all languages at once.
\textbf{Leave-one-out (LOO)}: a shared probe trained on all languages
except one and evaluated on the held-out language. This tests for language-agnostic semantic structure, since the probe has never seen the target language.
We also measure cross-lingual transfer with all pairwise combinations of single train--test languages, but we leave these results to Appendix~\ref{appendix:pairwise}.

\begin{figure*}[t!]
    \centering
    \includegraphics[width=1.0\textwidth]{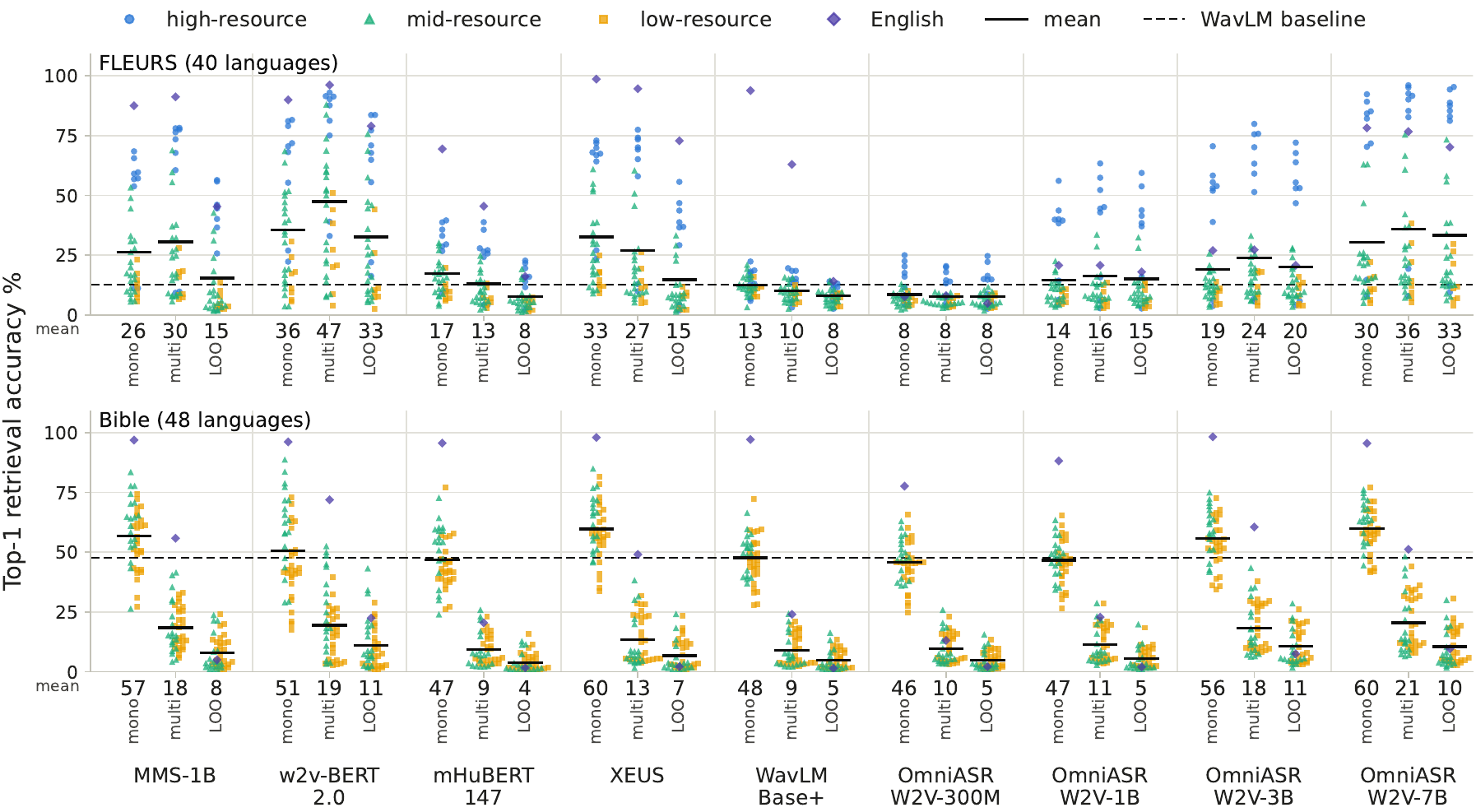}
    \caption{Monolingual, multilingual, and leave-one-out (LOO) top-1 accuracy of mean-pooled probes of self-supervised models on FLEURS (top) and Bible (bottom). Numbers along x-axis are means with English excluded.
    }
    \label{fig:probe-summary}
\end{figure*}

\paragraph{Layer Selection.}
We first sweep all layers with mean-pooled monolingual probes and select, per model, the layer that maximizes validation set retrieval on each corpus. All subsequent analyses use each model's selected layer, so models are compared at their respective best operating points.

\paragraph{Mean Centering.}
We experiment with per-language centering as an additional diagnostic: for each language, we
subtract the mean speech embedding computed over that language's
training set, before training and testing. The change in retrieval accuracy indicates
if some of the probe's performance depends on language- and speaker-specific directions.

\paragraph{Named Entity Control.}
One risk of our retrieval setup is that it reflects not sentence meaning but named entity shortcuts, such as names and places that sound alike across languages.
To test this risk, we prompt \texttt{gemma-4-31b-it} \citep{gemmateam2026gemma4technicalreport} to produce modified versions of every sentence: (i) \textbf{masked}, in which each entity is replaced by a generic placeholder of the same type (e.g., \emph{the place}, \emph{the organization}); and (ii) \textbf{names-only}, in which all non-entity content is stripped and only the entities remain. Masking tests whether entities are necessary (whether retrieval survives when names are removed). Names-only tests whether they are sufficient: if entities alone retrieve near chance, the signal is not carried by names.

\section{Results}

\begin{figure}[t!]
    \centering
    \includegraphics[width=1.0\columnwidth]{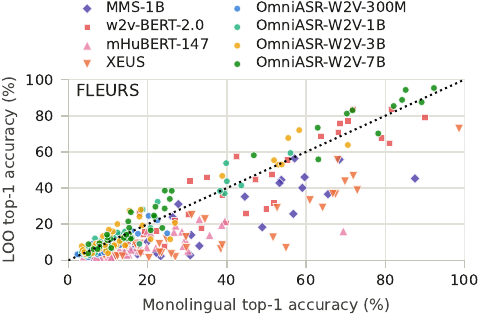}
    \caption{Monolingual against leave-one-out (LOO) retrieval accuracy for 40 FLEURS languages, showing the relationship between monolingual scores and cross-lingual transfer. OmniASR models show high transfer with many languages above the diagonal.}
    \label{fig:loo_vs_mono_scatter}
\end{figure}

\begin{table}[t!]
\centering
\small
\begin{tabular}{lrr}
\toprule
\textbf{Baseline} & \textbf{FLEURS} & \textbf{Bible} \\
\midrule
MFCC + $\Delta$ + $\Delta\Delta$ & 0.44 & 0.75 \\
Random WavLM (avg. 2 seeds)      & 0.38 & 1.08 \\
Duration                         & 0.63 & 0.10 \\
\bottomrule
\end{tabular}
\caption{Acoustic and duration baselines using monolingual top-1 accuracy (\%) averaged
over eight languages per corpus. MFCC baseline uses 40 MFCCs plus their first- and second-order deltas. Duration matches the $k$-th longest recording with $k$-th longest text candidate.}
\label{tab:acoustic-baselines}
\end{table}

\paragraph{Sentence meaning is linearly accessible but highly uneven across languages.}
While supervised speech and text encoders (such as SONAR) easily saturate the
retrieval task on FLEURS (known as $\operatorname{xsim}$) \citep{duquenne2023sonarsentencelevelmultimodallanguageagnostic},
this task remains difficult for self-supervised speech encoders.
The strongest models in our tests (w2v-BERT-2.0 and OmniASR-W2V-7B) achieve $>75\%$ monolingual accuracy (Fig.~\ref{fig:probe-summary})
purely from mean-pooled self-supervised features for several high-resource languages, but accuracy falls steeply on the mid- to low-resource tiers. WavLM achieves $13\%$ accuracy on non-English languages despite pre-training only on English, compared to $0.38\%$ randomly initialized (Table \ref{tab:acoustic-baselines}), indicating that a linear map can exploit pretrained acoustic features.

MMS-1B unexpectedly outperforms the $3\times$ larger
OmniASR-3B on FLEURS ($30\%$ vs.\ $24\%$ multi, winning on 29/39 non-English languages), despite sharing the
wav2vec~2.0 architecture and training on an order of magnitude fewer hours of audio.
We speculate that OmniASR allocates more of its training mixture across the long tail of languages, consistent with its weaker English scores.

Bible monolingual scores are consistently higher (e.g., XEUS $60\%$ vs.\ $33\%$ on FLEURS), with even the monolingual WavLM baseline surprisingly averaging $48\%$, indicating that single-speaker recordings allow probes to more easily exploit speaker- and channel-specific cues, inflating retrieval scores. We return to this below.

\begin{figure}[t!]
    \centering
    \includegraphics[width=1.0\columnwidth]{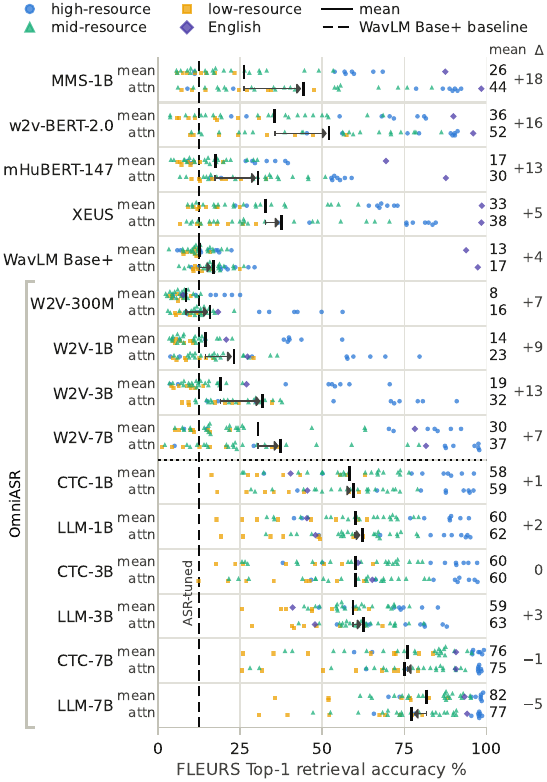}
    \caption{Monolingual retrieval accuracy on FLEURS, comparing mean- and attention-pooled probes. The higher scores of attention pooling on self-supervised models indicate temporally localized semantic content. OmniASR-300M ASR models excluded for brevity.}
    \label{fig:pooling}
\end{figure}

\begin{figure}[t!]
    \centering
    \includegraphics[width=1.0\columnwidth]{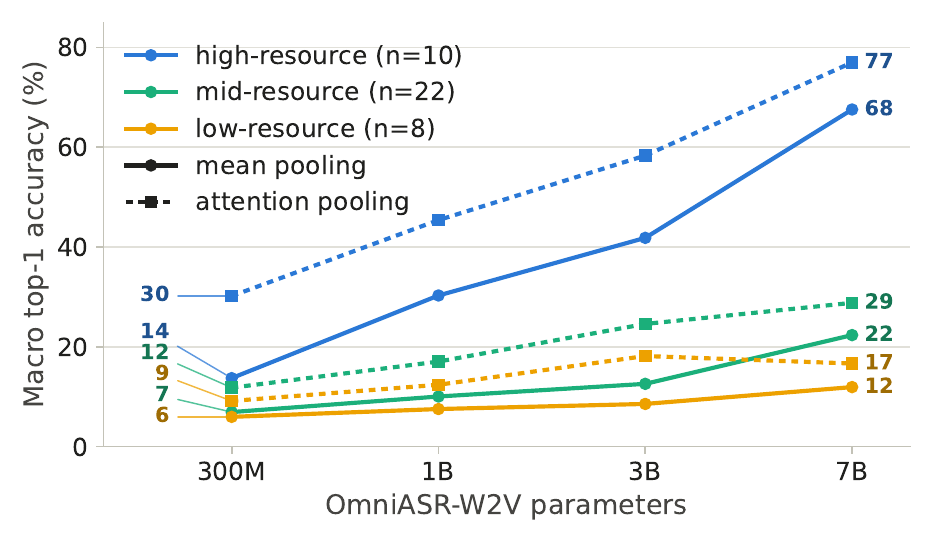}
    \caption{OmniASR-W2V top-1 retrieval on FLEURS at different model scales, broken down by language resource tier. Scaling the model up benefits high-resource languages more than low-resource languages.}
    \label{fig:omniasr_scaling}
\end{figure}

\paragraph{Cross-lingual semantic structure is present.}
On FLEURS, a shared, multilingual probe frequently matches or beats per-language
probes. For example, w2v-BERT-2.0's multilingual probe outperforms monolingual probes on 36/39 non-English languages ($47\%$ multi vs $36\%$ mono, Fig.~\ref{fig:probe-summary}), showing that one language-agnostic map can serve many languages.
Most models perform substantially worse in the LOO regime, with the exception of the OmniASR models (above the line in Figure~\ref{fig:loo_vs_mono_scatter}).
Remarkably, OmniASR-W2V-7B held-out wins on 24/39 languages vs. in-language ($33\%$ LOO vs.\ $30\%$ mono, Fig.~\ref{fig:probe-summary}), indicating broad multilingual gains that are not limited to a few languages.
However, LOO probes are trained on more utterances than monolingual probes, so this result shows that abundant supervision from other languages can compensate for scarce target-language data. More specifically, OmniASR-W2V-7B can exceed the monolingual mean without using the full LOO data, using around half of the full LOO data.\footnote{See Appendix \ref{appendix:loo-scaling} for LOO data scaling breakdown.}


\begin{figure*}[t!]
    \centering
    \includegraphics[width=1.0\textwidth]{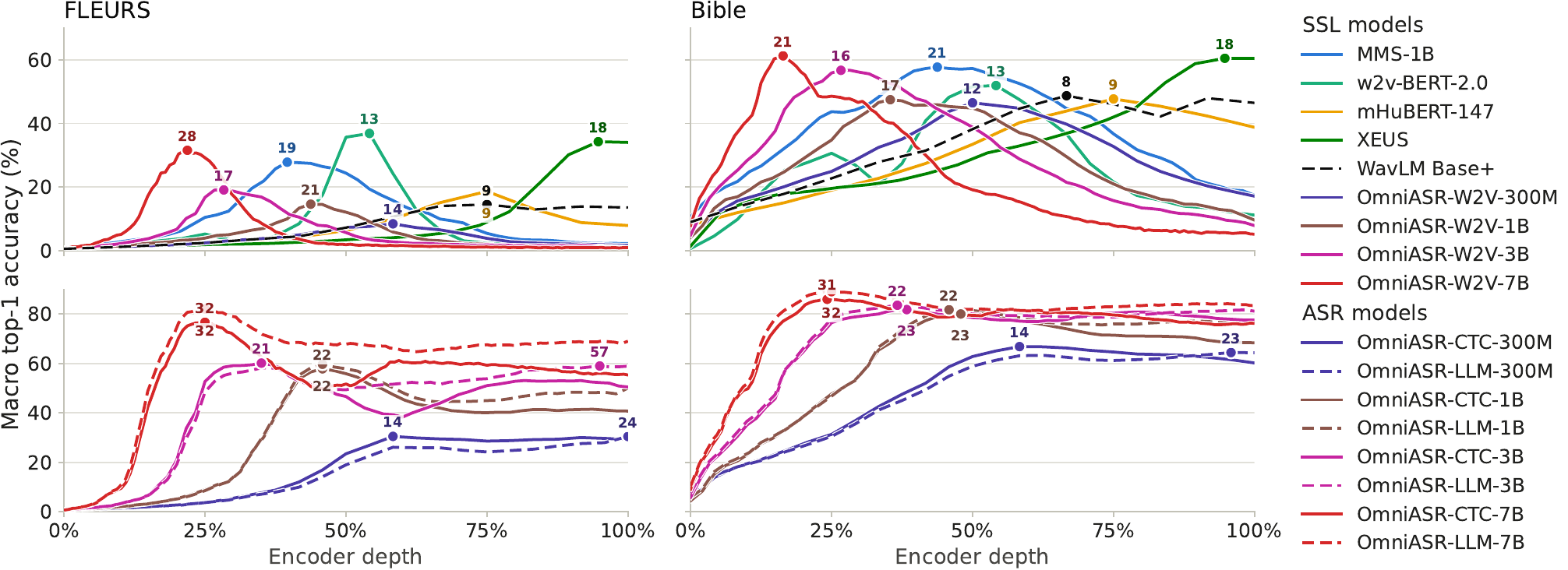}
    \caption{Layerwise monolingual top-1 accuracy on FLEURS (left) and Bible (right), comparing SSL (top) encoders and ASR-tuned models (bottom). Per-model peak layer indices are labeled. w2v-BERT-2.0, XEUS, and ASR models retain semantics in later layers compared to contrastive models (MMS, OmniASR-W2V).}
    \label{fig:layerwise}
\end{figure*}

\paragraph{Monolingual decodability does not predict cross-lingual transfer.}
XEUS has among the richest monolingual representations, on par with the much larger MMS-1B and OmniASR-W2V-3B models, yet transfers poorly ($33\%$ mono vs. $15\%$ LOO on FLEURS).
This pattern aligns with prior
work showing that the contrastive objective produces better
cross-lingual features for ASR compared to masked prediction \citep{li-etal-2023-quantitative}.

\paragraph{Scale isn't everything.}
Semantic quality is not a simple function of model scale:
w2v-BERT-2.0, at 580M parameters, has the strongest overall
FLEURS retrieval (47\% multi, Fig.~\ref{fig:probe-summary}) of
any self-supervised model, surpassing even OmniASR-W2V-7B, which is many times the size.
The most salient difference is its joint contrastive and masked-prediction objective,
but these models differ in many dimensions, including pretraining data, so further
work is needed to isolate the factor responsible for its stronger semantic decodability.

\paragraph{Semantic content is temporally localized.}
Attention pooling improves top-1 over mean pooling for every self-supervised model (Fig.~\ref{fig:pooling}). The gains are substantial: MMS rises from $26\%$ to $44\%$ and w2v-BERT-2.0 from $36\%$ to $52\%$, while ASR models are marginally affected, sometimes negatively. This indicates that self-supervised encoders do not spread semantic content evenly across frames, and mean pooling dilutes the semantic signal, whereas ASR tuning makes it more uniformly recoverable. However, since sweeping all probe regimes with attention pooling was
computationally prohibitive, our primary results use mean-pooled probes.

\paragraph{OmniASR-W2V scaling amplifies semantic disparities between resource tiers.}
Within the OmniASR-W2V family, scaling up from 300M$\rightarrow$7B produces sharply unequal gains across resource tiers (Fig.~\ref{fig:omniasr_scaling}). Monolingual probes rise $14\%$ to $68\%$ for the high-resource tier, compared with $7\%$ to $22\%$ for mid-resource and $6\%$ to $12\%$ for low-resource. Increased scale greatly improves semantic representations where speech data is abundant, while the low end remains relatively flat, widening the gap between resource tiers.\footnote{See Appendix~\ref{appendix:resource_tiers} for breakdown of resource tiers.}

\paragraph{OmniASR-W2V scaling adds depth primarily after the semantic peak.}
Figure~\ref{fig:layerwise} shows sharply different layerwise profiles
across different model types. The wav2vec~2.0 contrastive models exhibit
autoencoder-like behavior reported in prior work \citep{DBLP:conf/icassp/PasadSL23}:
retrieval accuracy rises to a peak in early layers and then
decays, as the final layers reconstruct input-like features.
OmniASR also reveals a scaling pattern: as total depth grows 24$\rightarrow$128 layers, the semantic peak only shifts 14$\rightarrow$28, barely growing with scale. Most extra depth is spent after the peak and is not relevant for this task.
The masked-prediction (mHuBERT-147) and denoising (XEUS) encoders
instead concentrate semantic content more deeply in the network, consistent with prior work attributing changes in layerwise organization to iterative refinement and pre-training objective \citep{DBLP:conf/interspeech/HuoD25, kloots2026trackingemergencelinguisticstructure}.

\begin{figure}[t!]
    \centering
    \includegraphics[width=1.0\columnwidth]{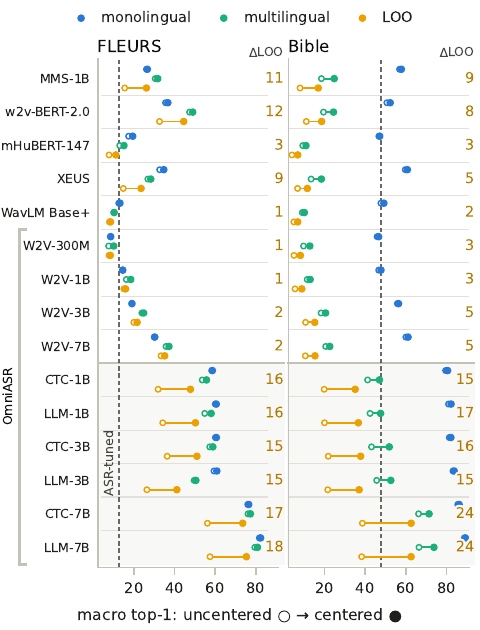}
    \caption{Mean (non-English) top-1 scores before and after mean centering. Dashed line marks the WavLM Base+ baseline. On FLEURS, centering closes the LOO vs. mono gap for w2v-BERT-2.0 and MMS-1B, and partially for ASR-tuned models, but has little effect on other models. On Bible, mean centering can partially close the gap but LOO still lags far behind mono results.}
    \label{fig:centering}
\end{figure}

\paragraph{ASR supervision strengthens semantic decodability, but in language-specific directions.}
Fine-tuning for transcription reshapes the self-supervised layerwise
profile, maintaining semantics in deeper layers and pushing the peak deeper (Fig.~\ref{fig:layerwise}).
It also sharply raises retrieval scores:
OmniASR-7B rises from 30\% (W2V) to 76\% (CTC) and
82\% (LLM) monolingual accuracy on FLEURS, with similar gains at
every scale for both CTC and LLM variants (Fig.~\ref{fig:pooling}).
However, relative to their self-supervised base models, the ASR-tuned models
show much larger mono--LOO gaps on FLEURS (e.g., OmniASR-7B LLM: 82\% mono
vs.\ 57\% LOO) and larger mean centering deltas
on both datasets (Fig.~\ref{fig:centering}), which indicates that ASR tuning strengthens semantic decodability in language-specific directions.

\paragraph{Semantic information is entangled with acoustics.}
On the Bible corpus, monolingual probes on self-supervised models can exceed the WavLM baseline while multilingual and LOO probes fall far short. This pattern is unlike multi-speaker FLEURS, where multilingual and LOO probes exceed the baseline on most models, suggesting that single-speaker probes benefit from speaker- and channel-specific acoustic regularities in the pretrained representations that do not transfer to new languages and speakers, thus inflating apparent semantic decodability.
Consistent with greater reliance on acoustic information, Bible also retains accuracy in later, more input-like layers of contrastive models (Fig.~\ref{fig:layerwise}). ASR tuning brings the OmniASR-7B LOO Bible probe above the baseline, showing that lexical supervision partially disentangles acoustic and semantic information.

Mean centering yields small improvements, especially for LOO probes, bringing LOO probes closer to multilingual probes in performance (Fig.~\ref{fig:centering}). However, it is not nearly enough to close the monolingual--LOO gap on the Bible corpus; this gap cannot be explained by a simple speaker- or language-specific offset.

\begin{figure}[t!]
    \centering
    \includegraphics[width=1.0\columnwidth]{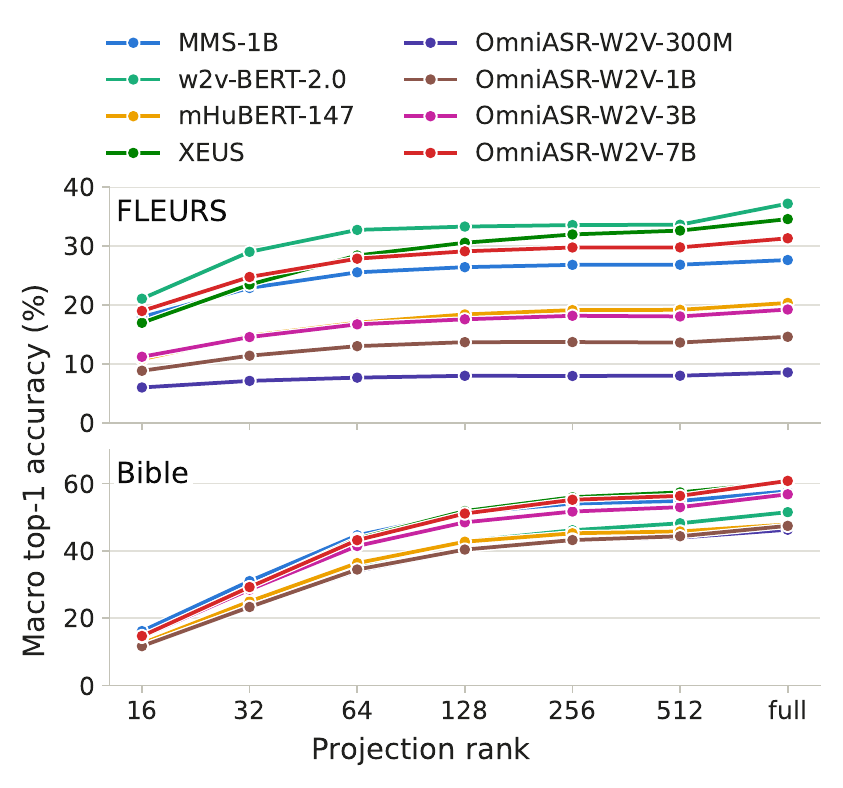}
    \caption{Mean monolingual top-1 as a function of projection rank for FLEURS (top) and Bible (bottom).
    }
    \label{fig:rank_scaling_fleurs_bible}
\end{figure}

\paragraph{Semantic decodability survives at low rank.}
We re-run the mean-pooled probes under
rank restriction (Fig.~\ref{fig:rank_scaling_fleurs_bible}).
Below rank 64, retrieval is weakened substantially, but rank~64
preserves most performance, and remains well above full-rank WavLM on FLEURS. Notably,
Bible probes benefit more from higher ranks, possibly by
using extra capacity to exploit many sparse, speaker-specific directions
on single-speaker training data.

\begin{figure}[t!]
    \centering
    \includegraphics[width=1.0\columnwidth]{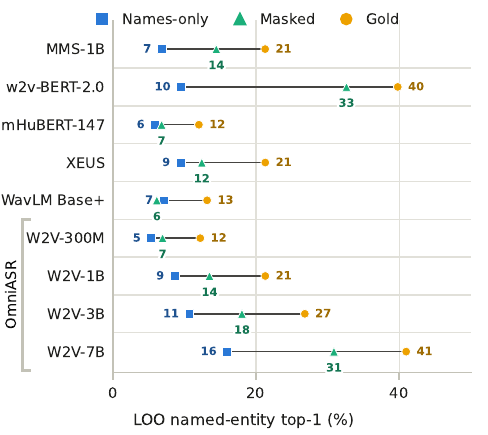}
    \caption{Named-entity results: leave-one-out (LOO) top-1 average (non-English) on FLEURS rows containing tagged entities.
    The strongest models retain substantial accuracy on \emph{Masked} sentences without entity information.
    \emph{Names-only} scores sit below \emph{Gold} scores, so names alone do not explain the strongest performance.
    }
    \label{fig:ne_loo}
\end{figure}

\paragraph{Semantic decodability is not merely an entity-matching shortcut.}
Masking named entities affects models unevenly (Fig.~\ref{fig:ne_loo}). Some reduction is expected, as masking removes legitimate information from the candidate sentences and creates a mismatch between the speech and target text. Therefore, we focus on the margin over the masked WavLM baseline rather than the size of the drop alone. For weaker self-supervised encoders, masked LOO accuracy falls to $7$–$14\%$, near the masked WavLM baseline of $6\%$, indicating that entities contribute materially when the semantic signal is weak. In contrast, w2v-BERT-2.0 and OmniASR-W2V-7B retain $33\%$ and $31\%$ accuracy after masking, respectively, well above the baseline.
Names-only retrieval is well below gold retrieval, so entities alone cannot explain the strongest results.
Thus, entities contribute information, but differences between the strongest and weakest encoders cannot be attributed to entity matching alone.

\section{Conclusion}

While far weaker than supervised encoders, S3Ms have
substantial decodable sentence-level semantic content for mid- to high-resource languages and, in some models, surprisingly cross-lingual representations. Yet cross-lingual capability is not predicted by monolingual performance: XEUS encodes rich within-language semantics but transfers poorly, while OmniASR models transfer well despite being semantically weak for their size.
Factors beyond sheer scale, such as training objective, impact semantic decodability, as the 580M-parameter w2v-BERT-2.0 surpasses the 7B OmniASR encoder. Our results indicate that supervised speech--text alignment can leverage semantic structure already present. However, we find semantic content entangled with acoustics that inflate apparent decodability on single-speaker corpora. We leave open how much of the residual gap to supervised systems reflects missing semantic content or content that is present but not linearly accessible.

\section*{Limitations}
Our analysis measures what is linearly decodable from frozen representations, not what a nonlinear trained system could extract, so our results could understate the semantic content actually present.
Conversely, due to acoustic associations, our results could overstate the semantic content;
low-rank and named-entity controls show that the effect does not require a full-capacity projection or entity-only matching, though other lexical cues may still contribute. Retrieval against a single frozen text encoder (SONAR) also means our results are conditioned on that encoder's own capabilities and biases; scores are strongly correlated with alternative text spaces (Appendix~\ref{sec:text-spaces}), but we did not repeat the full probing evaluation. Our two corpora, while complementary, are narrow in domain and register (Wikipedia-style news/geography and New Testament), and FLEURS lacks speaker IDs, preventing a clean separation of speaker confounds within that corpus. Language coverage, while broad relative to prior work, is still a small fraction of the full set of languages used to train these massively multilingual speech models.

\section*{Acknowledgments}
This work was supported by the ETEN Innovation Lab. We are grateful to Joel Mathew for his guidance and support during this project, and to the members of the OSU SLaTe Lab for helpful discussions. We also thank the anonymous reviewers for their valuable feedback and suggestions, which helped improve the quality of this paper.

\FloatBarrier

\raggedbottom
\bibliography{custom,anthology-1}
\clearpage
\flushbottom

\appendix

\onecolumn
\section{Language Resource Tiers}
\label{appendix:resource_tiers}

\newcommand{\probecorpus}[1]{%
  \addlinespace[2pt]\multicolumn{4}{@{}c@{}}{\textbf{#1}}\\[1pt]}
\newcommand{\probesource}[1]{%
  \addlinespace[3pt]\multicolumn{4}{@{}l@{}}{\textbf{#1}}\\[1pt]}
\newcommand{\probetier}[1]{%
  \addlinespace[2pt]\multicolumn{4}{@{}l@{}}{\textit{#1}}\\[1pt]}

\begin{table}[H]
\centering
\scriptsize
\setlength{\tabcolsep}{0pt}
\renewcommand{\arraystretch}{1.05}
\begin{tabular}{@{}
  >{\ttfamily}l @{\hspace{6pt}} l
  @{\hspace{20pt}}
  >{\ttfamily}l @{\hspace{6pt}} l @{}}
\toprule
\textbf{\normalfont Code} & \textbf{Language} &
\textbf{\normalfont Code} & \textbf{Language} \\
\midrule
\probecorpus{FLEURS}
\midrule
\probetier{High resource}
nl\_nl & Dutch (Netherlands) &
en\_us & English (United States) \\
fr\_fr & French (France) &
de\_de & German (Germany) \\
it\_it & Italian (Italy) &
pt\_br & Portuguese (Brazil) \\
es\_419 & Spanish (Latin America) &
cmn\_hans\_cn & Mandarin Chinese (Simpl., China) \\
ko\_kr & Korean (South Korea) &
ru\_ru & Russian (Russia) \\
\probetier{Mid resource}
pl\_pl & Polish (Poland) &
ar\_eg & Arabic (Egypt) \\
bn\_in & Bengali (India) &
el\_gr & Greek (Greece) \\
ha\_ng & Hausa (Nigeria) &
ja\_jp & Japanese (Japan) \\
fa\_ir & Persian (Iran) &
so\_so & Somali (Somalia) \\
sw\_ke & Swahili (Kenya) &
th\_th & Thai (Thailand) \\
tr\_tr & Turkish (Turkey) &
uk\_ua & Ukrainian (Ukraine) \\
ta\_in & Tamil (India) &
yo\_ng & Yoruba (Nigeria) \\
my\_mm & Burmese (Myanmar) &
hi\_in & Hindi (India) \\
id\_id & Indonesian (Indonesia) &
ig\_ng & Igbo (Nigeria) \\
te\_in & Telugu (India) &
ur\_pk & Urdu (Pakistan) \\
vi\_vn & Vietnamese (Vietnam) &
zu\_za & Zulu (South Africa) \\
\probetier{Low resource}
he\_il & Hebrew (Israel) &
fil\_ph & Filipino (Philippines) \\
am\_et & Amharic (Ethiopia) &
yue\_hant\_hk & Cantonese (Trad., Hong Kong) \\
ff\_sn & Fula (Senegal) &
km\_kh & Khmer (Cambodia) \\
lo\_la & Lao (Laos) &
mi\_nz & M\={a}ori (New Zealand) \\
\midrule
\probecorpus{Bible}
\midrule
\probesource{English WEB}
\probetier{High resource}
eng & English & & \\
\probesource{OpenBibleTTS}
\probetier{Mid resource}
arb & Modern Standard Arabic &
ben & Bengali \\
hau & Hausa &
kan & Kannada \\
swh & Swahili &
tam & Tamil \\
tur & Turkish &
yor & Yoruba \\
luo & Dholuo &
guj & Gujarati \\
hin & Hindi &
ibo & Igbo \\
kik & Kikuyu &
mal & Malayalam \\
pan & Punjabi &
sna & Shona \\
tel & Telugu &
urd & Urdu \\
vie & Vietnamese & & \\
\probetier{Low resource}
asm & Assamese &
ckb & Central Kurdish \\
hat & Haitian Creole &
lin & Lingala \\
lug & Luganda &
mar & Marathi \\
hne & Chhattisgarhi &
nya & Chichewa \\
dwr & Dawro &
ewe & Ewe \\
gmv & Gamo &
gof & Gofa \\
hil & Hiligaynon &
nde & North Ndebele \\
nep & Nepali &
orm & Oromo \\
twi\_akuapem & Twi (Akuapem) &
twi\_asante & Twi (Asante) \\
\probesource{Snow Mountain}
\probetier{Low resource}
bhadrawahi & Bhadrawahi &
bilaspuri & Bilaspuri \\
dogri & Dogri &
gaddi & Gaddi \\
haryanvi & Haryanvi &
kangri & Kangri \\
kulvi & Kulvi &
kulvi\_outer\_seraji & Kulvi Outer Seraji \\
mandeali & Mandeali &
pahari\_mahasui & Pahari Mahasui \\
\bottomrule
\end{tabular}
\caption{Probe languages grouped by dataset and resource tier. Tiers use
the per-language hours in \textsc{AllASR} published with Omnilingual ASR: high ($\geq 2{,}000$h), mid
($200 \leq h < 2{,}000$), and low ($<200$h or missing).
These tiers serve as a rough proxy for broad speech resource availability.
Language codes are listed as provided by respective datasets.
Languages occurring in both corpora are listed
in both sections, giving 88 language entries and 76 distinct languages.
Twelve languages (en, hi, ur, bn, ta, te, tr, vi, sw, ha, yo, ig) are shared between the two corpora.
}
\label{tab:probe-languages}
\end{table}




\clearpage
\onecolumn
\section{Probe Seed Variability}
\label{appendix:probe_seed}

\begin{table}[H]
  \centering
  \setlength{\tabcolsep}{8pt}
  \small
  \begin{tabular}{@{}lrcc@{}}
    \toprule
    \textbf{Probe Type} &
    \textbf{Probe Count} &
    \textbf{Top-1 $\uparrow$} &
    \textbf{Top-1 Range} \\
    \midrule
    \multicolumn{4}{@{}l}{\textbf{w2v-BERT-2.0} (layer 13)} \\
    \addlinespace[1pt]
    Monolingual         & 30 & $44.85 \pm 0.35$ & $[44.24,\,45.11]$ \\
    Multilingual        &  5 & $51.11 \pm 0.43$ & $[50.75,\,51.74]$ \\
    LOO \hspace{4cm}    & 30 & $21.88 \pm 0.48$ & $[21.12,\,22.23]$ \\
    \addlinespace[3pt]
    \multicolumn{4}{@{}l}{\textbf{OmniASR-W2V-7B} (layer 28)} \\
    \addlinespace[1pt]
    Monolingual       & 30 & $39.05 \pm 0.09$ & $[38.93,\,39.19]$ \\
    Multilingual      &  5 & $41.62 \pm 0.34$ & $[41.30,\,42.11]$ \\
    LOO  & 30 & $27.19 \pm 0.10$ & $[27.08,\,27.35]$ \\
    \bottomrule
  \end{tabular}
  \caption{Probe seed variability on FLEURS. Results are macro-averaged over
  six languages (\texttt{en\_us}, \texttt{de\_de}, \texttt{hi\_in},
  \texttt{cmn\_hans\_cn}, \texttt{sw\_ke}, and \texttt{yo\_ng}) across five seeds.
  Probe count is the number of individually trained probes:
  six per seed for monolingual and leave-one-out (LOO) and one per seed for
  multilingual. The Top-1 column shows the mean $\pm$ sample standard
  deviation across the five seed-level language macro-averages; the range gives
  the observed minimum and maximum top-1. Probe
  layers and all other experimental settings are fixed across seeds.}
  \label{tab:probe-seed-variability}
\end{table}

\section{LOO Data Scaling}
\label{appendix:loo-scaling}

\begin{table}[H]
    \centering
    \begin{minipage}[t]{0.48\textwidth}
        \centering
        \textbf{(a) Adding full languages}\\[4pt]
        \begin{tabular}{@{}rrrr@{}}
            \toprule
            \textbf{Languages ($k$)} & \textbf{Utterances} & \textbf{Top-1 (\%)} & \textbf{Wins} \\
            \midrule
             1 &   2.7k &  6.4 &  0/39 \\
             2 &   5.3k & 11.7 &  0/39 \\
             4 &  10.6k & 17.6 &  0/39 \\
             8 &  21.0k & 24.0 &  3/39 \\
            16 &  42.2k & 29.9 & 16/39 \\
            24 &  63.5k & \textbf{31.6} & 23/39 \\
            32 &  84.7k & \textbf{33.0} & 25/39 \\
            39 & 103.2k & \textbf{33.2} & 27/39 \\
            \bottomrule
        \end{tabular}
    \end{minipage}\hfill
    \begin{minipage}[t]{0.48\textwidth}
        \centering
        \textbf{(b) Scaling data with all languages}\\[4pt]
        \begin{tabular}{@{}rrrr@{}}
            \toprule
            \textbf{Languages} & \textbf{Utterances} & \textbf{Top-1 (\%)} & \textbf{Wins} \\
            \midrule
            39 &   2.6k & 14.8 &  0/39 \\
            39 &   5.2k & 19.6 &  0/39 \\
            39 &  10.4k & 24.4 &  3/39 \\
            39 &  26.0k & 29.5 & 15/39 \\
            39 &  52.0k & \textbf{31.7} & 23/39 \\
            39 & 103.2k & \textbf{33.2} & 27/39 \\
            \bottomrule
        \end{tabular}
    \end{minipage}
    \caption{
        LOO probe training data scaling behavior for OmniASR-W2V-7B on FLEURS.
        Each LOO probe is evaluated on one held-out language, and
        Top-1 is the mean over all 39 non-English held-out languages.
        Wins counts the held-out languages on which the LOO probe
        matches or exceeds that language's monolingual probe.
        \textbf{(a)} Each probe is trained on $k$ other languages, each
        contributing its full train split, so language diversity and number of utterances grows with $k$. The probe's score is averaged over three random draws of $k$ languages.
        \textbf{(b)} All 39 source languages are retained while varying
        the number of utterances via subsampling. Scores are averaged over three random draws.
        Scores exceeding the monolingual
        mean of 30.4\% are bolded.
    }
    \label{tab:loo-scaling}
\end{table}

\section{Mean Centering by Resource Tier}
\label{appendix:centering_tiers}

\begin{table}[H]
  \centering
  \setlength{\tabcolsep}{5pt}
  \begin{tabular}{@{}lccccc@{}}
    \toprule
    & \multicolumn{3}{c}{\textbf{FLEURS}} & \multicolumn{2}{c}{\textbf{Bible}} \\
    \cmidrule(lr){2-4}\cmidrule(l){5-6}
    \textbf{Models} & High & Mid & Low & Mid & Low \\
    \midrule
    Other SSL (4)         & 27.5 $\to$ 39.2 & 18.1 $\to$ 28.1 & \phantom{0}7.0 $\to$ 11.7 & 12.1 $\to$ 20.8 & \phantom{0}6.2 $\to$ 10.8 \\
    OmniASR-W2V (4)       & 29.3 $\to$ 30.7 & 15.8 $\to$ 17.3 & \phantom{0}8.7 $\to$ \phantom{0}9.6 & 12.0 $\to$ 16.8 & \phantom{0}7.1 $\to$ 10.3 \\
    OmniASR ASR-tuned (8) & 41.9 $\to$ 54.9 & 37.2 $\to$ 51.6 & 17.3 $\to$ 31.4 & 35.1 $\to$ 52.6 & 19.6 $\to$ 33.2 \\
    \bottomrule
  \end{tabular}
  \caption{LOO top-1 accuracy by resource tier
  (Appendix~\ref{appendix:resource_tiers}), uncentered $\to$ mean-centered,
  averaged over models in each group (Other SSL: mHuBERT-147, XEUS,
  w2v-BERT-2.0, MMS-1B). Centering improves every tier, and the low-resource
  tier remains the weakest after centering. The Bible high
  tier (English only) is omitted.}
  \label{tab:centering-tiers}
\end{table}

\clearpage

\section{Per-language Probe Results}
\label{appendix:per-language}


\begin{figure}[H]
    \centering
    \includegraphics[width=1.0\textwidth]{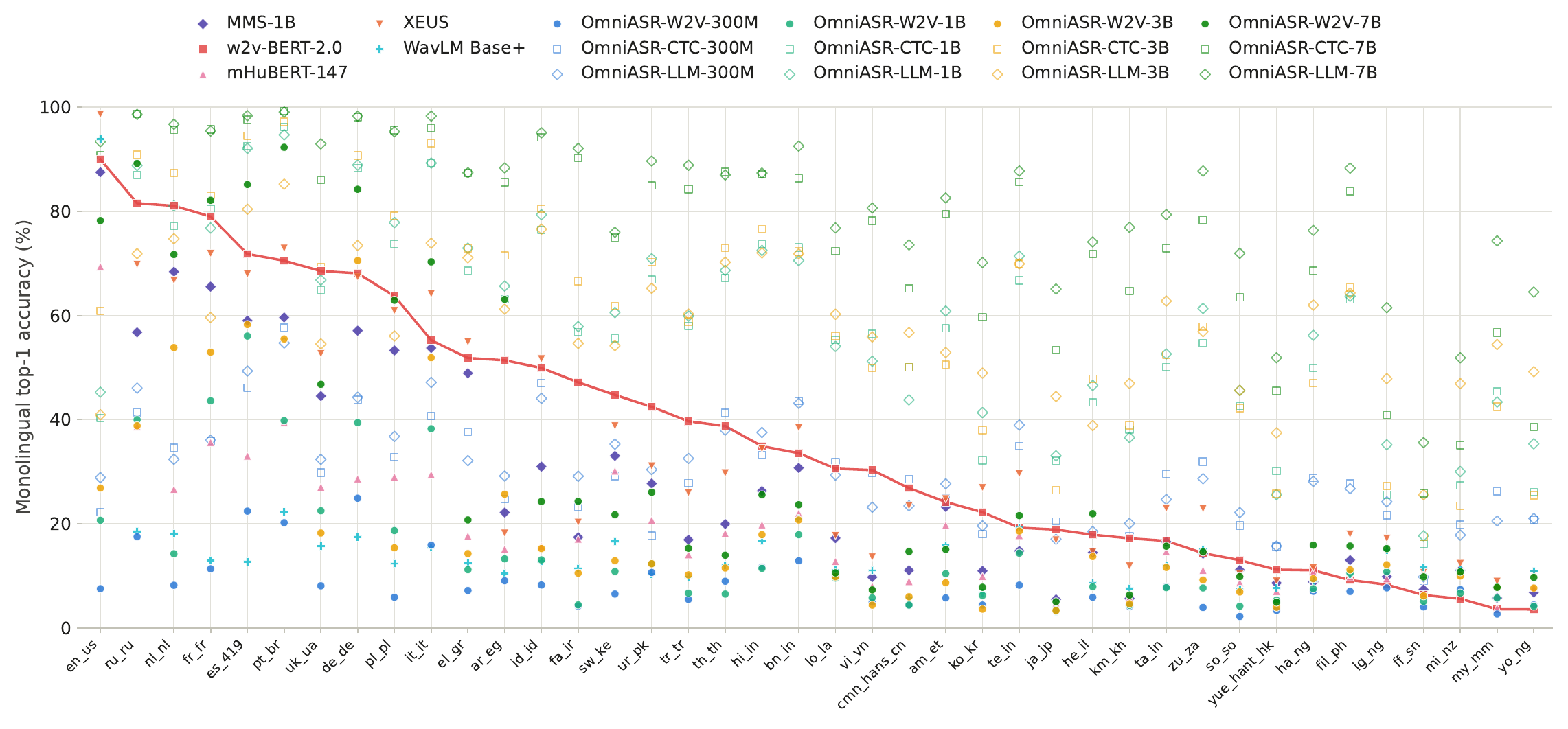}
    \caption{Monolingual probe retrieval accuracy on the FLEURS test set for 40 languages. The red line connects the points for w2v-BERT-2.0, the self-supervised model with best average monolingual probe scores on this test set. ASR-tuned models (OmniASR CTC and LLM variants) have much higher scores due to text supervision.}
    \label{fig:per_lang_fleurs}
\end{figure}

\begin{figure}[H]
    \centering
    \includegraphics[width=1.0\textwidth]{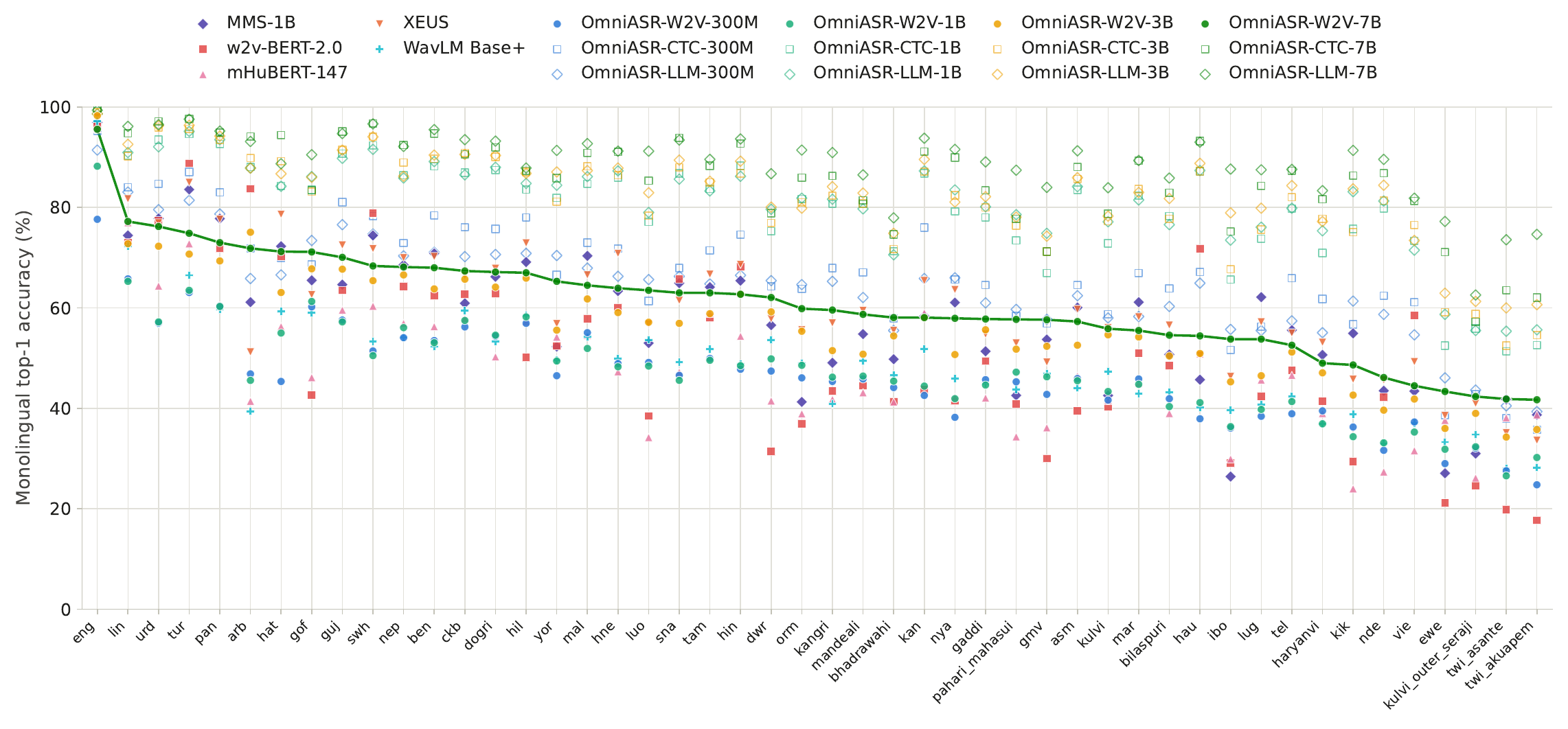}
    \caption{Monolingual probe retrieval accuracy on the Bible test set for 48 languages. The green line connects the points for OmniASR-W2V-7B, the self-supervised model with best average monolingual probe scores on this test set. ASR-tuned models (OmniASR CTC and LLM variants) have much higher scores due to text supervision.}
    \label{fig:per_lang_bible}
\end{figure}



\clearpage
\onecolumn
\section{Pairwise Probe Evaluation}
\label{appendix:pairwise}
\begin{figure}[H]
    \centering
    \includegraphics[width=1.0\textwidth]{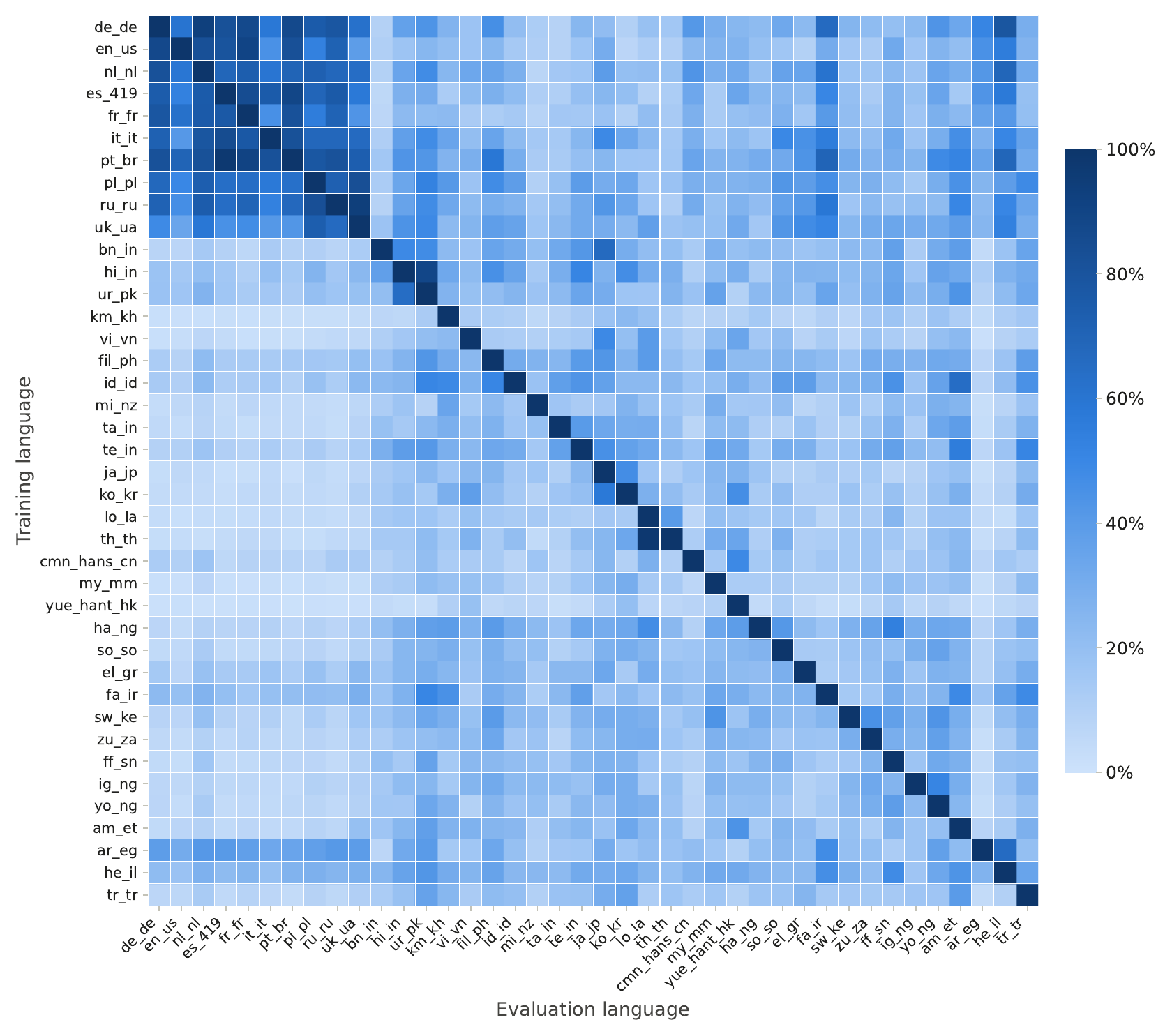}
    \caption{Pairwise cross-lingual transfer using OmniASR-W2V-7B on FLEURS, using mean-pooled probes trained on one language and tested on another. Pairwise transfer is calculated as the top-1 for a given train-test language pair divided by the monolingual probe's top-1 on that language, measuring the ratio of monolingual performance that is maintained by zero-shot cross-lingual transfer. Pairwise transfer often benefits related language sources more than unrelated ones (e.g., Hindi--Urdu and Lao--Thai have high transfer, while English--Hindi has poor transfer). However, high-resource languages transfer better in general compared to low-resource languages regardless of relatedness.}
    \label{fig:full_pairwise}
\end{figure}

\twocolumn
\clearpage

\section{Named-Entity Controls}
\label{sec:ne-prompts}

\raggedbottom
One risk of our retrieval setup is that it reflects not sentence meaning but named entity shortcuts, such as names and places that sound alike across languages. Our Bible corpus additionally has the risk of common usage of names that are transliterated rather than translated, causing a high degree of similarity across languages.

We test if this type of risk applies in the speech domain by generating three controlled variants of each English target text (only the first two are used in our main analysis): (i) \textbf{masked}, in which each entity is replaced by a generic placeholder of the same type (e.g., \emph{the place}, \emph{the organization}, \emph{the man}); (ii) \textbf{names-only}, in which all non-entity content is stripped and only the entities remain; and (iii) \textbf{swapped}, in which each entity is replaced by a different, type-consistent entity, yielding a similar sentence with different referents. Masking tests whether entities are necessary (whether retrieval survives when names are removed). Names-only tests whether they are sufficient: if entities alone retrieve near chance, the signal is not carried by names. The swap test measures entity discrimination: whether the speech embedding prefers the gold text over the entity-swapped version, reported as a paired win rate against the chance value of $0.5$.

\subsection{Prompt Construction}
The named-entity controls are generated by prompting \texttt{gemma-4-31B-it} \citep{gemmateam2026gemma4technicalreport} per English
target text. The prompt was drafted initially by Claude Opus 4.8 and manually adjusted.
The prompting procedure uses a JSON object \texttt{\{"text": "..."\}} as input and
a JSON object as output with the tagged entities and the three derived sentences.
We use greedy decoding, with retries (temperature~0.7) for rows whose
first completion fails JSON parsing or consistency checks (e.g., the
\emph{names-masked} text still containing a tagged surface form, or the
\emph{name-swapped} text being identical to the input). Because the two corpora
differ in domain and in how proper names behave, we use separate prompts for each:
the FLEURS prompt targets Wikipedia-style topics, while the Bible prompt
is specialized to New Testament verses.
Table~\ref{tab:ne-variant-example} shows the results of this pipeline with a FLEURS training row with all
three generated variants. Table~\ref{tab:ne-control} contains the complete retrieval results for both FLEURS and Bible, for both self-supervised and ASR-tuned models.

\subsection{FLEURS Prompt}

\begin{Verbatim}
# Task

You are generating controlled text variants for a speech–text retrieval probing experiment. The input is an English sentence that has a spoken translation in another language. We need to measure how much retrieval relies on proper names (which often sound similar across languages). You will (1) identify proper-name entities and (2) produce three derived texts: a name-swapped twin, a names-masked version, and a names-only string.

Work carefully and deterministically. Do not add commentary.

## Input

A single JSON object: `{"text": "..."}`

## Output

A single JSON object, nothing else:

```json
{
  "name_bearing": true|false,
  "entities": [{"surface": "...", "type": "PERSON|PLACE|ORG|GROUP|WORK|EVENT", "gender": "m|f|null"}],
  "replacements": [{"original": "...", "replacement": "...", "type": "..."}],
  "name_swapped": "..." | null,
  "names_masked": "..." | null,
  "names_only": "..." | null
}
```

## Entity tagging rules

- Tag only proper names: PERSON (people), PLACE (cities, countries, regions, rivers, mountains, landmarks), ORG (companies, institutions, agencies, teams), GROUP (nationalities, ethnic or religious groups used as proper nouns), WORK (titles of books, films, works), EVENT (named events, wars, festivals).
- Do NOT tag: numbers, dates, months, weekdays, units, common nouns, job titles or generic role words, languages used adjectivally in ordinary description.
- List each distinct entity once in `entities`, even if mentioned multiple times.
- Set `gender` for PERSON entities when inferable (needed for pronoun consistency), otherwise null.

## name_swapped rules

Produce a twin sentence identical to `text` except that EVERY entity is replaced:

- Replace each entity with a different real entity of the same type. Replace all mentions of the same entity with the same replacement.
- Leave every other word, all punctuation, and all casing untouched. Adjust only obligatory grammar (e.g., "a"/"an").
- The replacement must be acoustically dissimilar from the original: different initial sound, no long shared substring. Never use a spelling or cognate variant (John→Jon, Maria→Marie, Brazil→Brasil are all forbidden).
- The replacement must not already appear anywhere in the sentence, and must not be acoustically similar to any other word in the sentence.
- For PERSON entities that are coreferent with pronouns in the sentence, match gender.
- The result must be fluent and type-plausible (a person where a person was, a city where a city was). It is fine — expected — that the twin states something factually false; only grammaticality and type-plausibility matter.

## names_masked rules

Replace each entity mention with a generic phrase of the same type that fits the syntax: PERSON→"the man"/"the woman"/"the person", PLACE→"the place"/"the region"/"the city", ORG→"the organization", GROUP→"the people", WORK→"the work", EVENT→"the event". Do not paraphrase or alter anything outside the entity spans, but the result MUST be grammatical English:

- Merge the placeholder with surrounding determiners, titles, and modifiers so no article clash occurs. Never output "the the", "an the", or a title directly before a placeholder: "President Obama"→"the president", "Dr. Smith"→"the doctor", "Ancient China"→"that ancient land", "Mount Everest"→"the mountain" — the title or modifier absorbs into the placeholder instead of preceding it.
- Adjacent entity mentions that refer to one thing become ONE placeholder, not two.
- Capitalize the placeholder when it starts a sentence ("The man, called to be...").

## names_only rules

The distinct entity surface forms in order of first appearance, joined by "; ". Example: "Paul; Macedonia".

## Sentences without entities

Set `name_bearing` to false, `entities` and `replacements` to `[]`, and `name_swapped`, `names_masked`, `names_only` to null.

# Domain notes (FLEURS)

- Sentences come from English Wikipedia-style text: news, geography, science, sport, travel.
- Sample replacements from real-world entities of the same type and roughly comparable prominence (a world capital for a world capital, a national politician for a national politician, a well-known company for a well-known company).
- Acronym entities (NASA, FIFA, EU) count as ORG; replace with another acronym or short-name organization of similar kind.
- Demonyms and country adjectives ("French wine", "Brazilian team") count as GROUP/PLACE usage — tag and swap them, keeping the adjectival form ("Turkish wine", "Peruvian team").
- Product and brand names count as ORG.
\end{Verbatim}

\subsection{Bible Prompt}

\begin{Verbatim}
# Task

You are generating controlled text variants for a speech–text retrieval probing experiment. The input is an English New Testament verse that has spoken translations in many languages. We need to measure how much retrieval relies on proper names (which are transliterated, and therefore sound similar, across most Bible translations). You will (1) identify proper-name entities and (2) produce three derived texts: a name-swapped twin, a names-masked version, and a names-only string.

Work carefully and deterministically. Do not add commentary.

## Input

A single JSON object: `{"text": "..."}`

## Output

A single JSON object, nothing else:

```json
{
  "name_bearing": true|false,
  "entities": [{"surface": "...", "type": "PERSON|PLACE|ORG|GROUP|WORK|EVENT", "gender": "m|f|null"}],
  "replacements": [{"original": "...", "replacement": "...", "type": "..."}],
  "name_swapped": "..." | null,
  "names_masked": "..." | null,
  "names_only": "..." | null
}
```

## Entity tagging rules

- Tag only proper names: PERSON (people), PLACE (cities, regions, rivers, mountains), ORG, GROUP (peoples and sects used as proper nouns), WORK, EVENT.
- Do NOT tag: numbers, dates, common nouns, generic role words ("the centurion", "a prophet"), or translated titles (see domain notes).
- List each distinct entity once in `entities`, even if mentioned multiple times.
- In multi-part references like "Simon son of Jonah", tag "Simon" and "Jonah" as separate PERSON entities.
- Set `gender` for PERSON entities when inferable (needed for pronoun consistency), otherwise null.

## name_swapped rules

Produce a twin verse identical to `text` except that EVERY entity is replaced:

- Replace each entity with a different entity of the same type. Replace all mentions of the same entity with the same replacement.
- Leave every other word, all punctuation, and all casing untouched. Adjust only obligatory grammar (e.g., "a"/"an").
- The replacement must be acoustically dissimilar from the original: different initial sound, no long shared substring. Never use a spelling or cognate variant (Simon→Simeon, Judas→Jude, Mary→Miriam are all forbidden).
- The replacement must not already appear anywhere in the verse, and must not be acoustically similar to any other word in the verse.
- For PERSON entities that are coreferent with pronouns in the verse, match gender.
- The result must be fluent and type-plausible (a person where a person was, a city where a city was). It is fine — expected — that the twin states something factually false; only grammaticality and type-plausibility matter.

## names_masked rules

Replace each entity mention with a generic phrase of the same type that fits the syntax: PERSON→"the man"/"the woman", PLACE→"the place"/"the city"/"the region", GROUP→"the people", ORG→"the assembly". Do not paraphrase or alter anything outside the entity spans, but the result MUST be grammatical English:

- Merge the placeholder with surrounding determiners, titles, and modifiers so no article clash occurs. Never output "the the" or a title directly before a placeholder: "our Lord Jesus Christ"→"our Lord", "King Herod"→"the king", "Mount Sinai"→"the mountain" — the title absorbs the name instead of preceding a placeholder.
- Adjacent entity mentions that refer to one person become ONE placeholder, not two: "Christ Jesus"→"the man", never "the man the man".
- Capitalize the placeholder when it starts a sentence or verse ("The man, called to be an apostle...").

## names_only rules

The distinct entity surface forms in order of first appearance, joined by "; ". Example: "Paul; Macedonia".

## Verses without entities

Set `name_bearing` to false, `entities` and `replacements` to `[]`, and `name_swapped`, `names_masked`, `names_only` to null.

# Domain notes (New Testament)

- **Tag transliterated names; do not tag translated titles.** The experimental confound is acoustic sharing through transliteration, so the rule is: words rendered as names across Bible translations are entities; words rendered as translated titles are not.
  - TAG and swap: Jesus, Christ (acoustically name-like and transliterated), Peter, Paul, Mary, Pilate, Herod, Jerusalem, Galilee, Capernaum, Jordan, Pharisees, Sadducees, Samaritans, etc.
  - Do NOT tag: God, Lord, the Father, the Son, the Holy Spirit, the Spirit, the Most High, Son of Man, the Devil, heaven, the temple, the synagogue.
- Sample replacements from the same domain: swap people for other biblical people of similar narrative role and prominence (an apostle for an apostle, a ruler for a ruler), places for other biblical places of the same kind (a Galilean town for a Galilean town, a region for a region).
- Each verse is processed independently; replacements do NOT need to be consistent across verses.
\end{Verbatim}

\begin{table*}[t!]
\centering
\begin{tabular}{>{\raggedright\arraybackslash}p{0.12\textwidth} >{\raggedright\arraybackslash}p{0.82\textwidth}}
\toprule
\textbf{Variant} & \textbf{Text} \\
\midrule
\textit{Original} & The pilot was identified as Squadron Leader Dilokrit Pattavee. \\
\textit{Masked} & The pilot was identified as the squadron leader. \\
\textit{Names-only} & Dilokrit Pattavee \\
\textit{Swapped} & The pilot was identified as Squadron Leader Marcus Thorne. \\
\bottomrule
\end{tabular}
\caption{Example FLEURS training row (ID 4) and its three generated named-entity control variants.}
\label{tab:ne-variant-example}
\end{table*}

\begin{table*}[t!]
\small
\setlength{\tabcolsep}{2.5pt}
\centering
\begin{tabular}{l@{\hspace{15pt}}
    *{4}{>{\raggedleft\arraybackslash}p{1.15em}}@{\hspace{15pt}}
    *{4}{>{\raggedleft\arraybackslash}p{1.15em}}@{\hspace{15pt}}
    *{4}{>{\raggedleft\arraybackslash}p{1.15em}}}
\toprule
& \multicolumn{4}{c}{\textbf{Mono.}} & \multicolumn{4}{c}{\textbf{Multi.}}
& \multicolumn{4}{c}{\textbf{LOO}} \\
\cmidrule(lr){2-5}\cmidrule(lr){6-9}\cmidrule(lr){10-13}
\textbf{Model}
& \multicolumn{1}{c}{G} & \multicolumn{1}{c}{M} & \multicolumn{1}{c}{N} & \multicolumn{1}{|c}{\emph{S}}
& \multicolumn{1}{c}{G} & \multicolumn{1}{c}{M} & \multicolumn{1}{c}{N} & \multicolumn{1}{|c}{\emph{S}}
& \multicolumn{1}{c}{G} & \multicolumn{1}{c}{M} & \multicolumn{1}{c}{N} & \multicolumn{1}{|c}{\emph{S}} \\
\midrule
\multicolumn{13}{l}{\textit{FLEURS}} \\
\addlinespace[2pt]
MMS-1B & 33 & 25 & 9 & 68 & 39 & 26 & 12 & 79 & 21 & 14 & 7 & 70 \\
w2v-BERT-2.0 & 42 & 35 & 9 & 66 & 55 & 46 & 13 & 76 & 40 & 33 & 10 & 71 \\
mHuBERT-147 & 23 & 18 & 7 & 68 & 20 & 11 & 8 & 75 & 12 & 7 & 6 & 71 \\
XEUS & 41 & 30 & 12 & 75 & 36 & 23 & 14 & 82 & 21 & 12 & 9 & 74 \\
WavLM Base+ & 18 & 13 & 6 & 67 & 17 & 8 & 9 & 77 & 13 & 6 & 7 & 75 \\
OmniASR-W2V-300M & 12 & 9 & 4 & 62 & 12 & 7 & 4 & 70 & 12 & 7 & 5 & 71 \\
OmniASR-W2V-1B & 19 & 14 & 6 & 65 & 22 & 15 & 8 & 73 & 21 & 14 & 9 & 74 \\
OmniASR-W2V-3B & 25 & 19 & 8 & 67 & 31 & 21 & 12 & 77 & 27 & 18 & 11 & 74 \\
OmniASR-W2V-7B & 36 & 29 & 11 & 69 & 44 & 33 & 17 & 78 & 41 & 31 & 16 & 78 \\
OmniASR-CTC-300M & 38 & 28 & 10 & 72 & 15 & 10 & 5 & 71 & 8 & 5 & 3 & 65 \\
OmniASR-LLM-300M & 40 & 29 & 11 & 74 & 33 & 18 & 12 & 81 & 20 & 11 & 8 & 75 \\
OmniASR-CTC-1B & 66 & 53 & 20 & 79 & 64 & 48 & 22 & 84 & 39 & 28 & 13 & 74 \\
OmniASR-LLM-1B & 68 & 55 & 20 & 80 & 66 & 49 & 22 & 85 & 43 & 31 & 16 & 77 \\
OmniASR-CTC-3B & 67 & 56 & 20 & 79 & 68 & 52 & 24 & 85 & 45 & 33 & 16 & 77 \\
OmniASR-LLM-3B & 70 & 51 & 26 & 86 & 64 & 39 & 31 & 88 & 39 & 20 & 20 & 83 \\
OmniASR-CTC-7B & 82 & 72 & 26 & 81 & 84 & 71 & 30 & 87 & 64 & 52 & 22 & 79 \\
OmniASR-LLM-7B & 87 & 75 & 34 & 88 & 88 & 73 & 37 & 91 & 67 & 51 & 26 & 83 \\
\midrule
\multicolumn{13}{l}{\textit{Bible}} \\
\addlinespace[2pt]
MMS-1B & 61 & 48 & 5 & 81 & 24 & 17 & 3 & 79 & 11 & 7 & 2 & 71 \\
w2v-BERT-2.0 & 54 & 45 & 3 & 77 & 25 & 18 & 2 & 77 & 15 & 11 & 2 & 72 \\
mHuBERT-147 & 51 & 41 & 3 & 77 & 14 & 9 & 1 & 79 & 6 & 4 & 1 & 72 \\
XEUS & 63 & 51 & 5 & 82 & 19 & 12 & 3 & 81 & 10 & 6 & 2 & 77 \\
WavLM Base+ & 51 & 41 & 3 & 77 & 13 & 8 & 2 & 80 & 8 & 5 & 1 & 75 \\
OmniASR-W2V-300M & 49 & 39 & 3 & 77 & 15 & 9 & 2 & 81 & 8 & 5 & 1 & 75 \\
OmniASR-W2V-1B & 50 & 40 & 3 & 78 & 17 & 10 & 2 & 80 & 8 & 5 & 2 & 76 \\
OmniASR-W2V-3B & 60 & 47 & 5 & 81 & 25 & 15 & 5 & 84 & 16 & 9 & 4 & 80 \\
OmniASR-W2V-7B & 64 & 50 & 6 & 81 & 28 & 17 & 5 & 84 & 16 & 8 & 4 & 80 \\
OmniASR-CTC-300M & 69 & 56 & 6 & 83 & 26 & 17 & 3 & 82 & 11 & 7 & 2 & 71 \\
OmniASR-LLM-300M & 66 & 53 & 5 & 83 & 22 & 14 & 3 & 80 & 12 & 8 & 2 & 75 \\
OmniASR-CTC-1B & 82 & 67 & 10 & 88 & 49 & 34 & 7 & 87 & 25 & 16 & 4 & 77 \\
OmniASR-LLM-1B & 83 & 68 & 10 & 88 & 51 & 35 & 7 & 87 & 25 & 16 & 4 & 78 \\
OmniASR-CTC-3B & 83 & 70 & 10 & 87 & 50 & 36 & 6 & 85 & 27 & 18 & 5 & 77 \\
OmniASR-LLM-3B & 85 & 71 & 11 & 88 & 54 & 37 & 8 & 86 & 27 & 17 & 5 & 77 \\
OmniASR-CTC-7B & 87 & 74 & 11 & 87 & 73 & 56 & 9 & 87 & 45 & 33 & 6 & 79 \\
OmniASR-LLM-7B & 90 & 77 & 13 & 89 & 73 & 56 & 10 & 87 & 45 & 32 & 7 & 80 \\
\bottomrule
\end{tabular}
\caption{Named-entity control results (\%) averaged over the non-English languages in FLEURS and Bible rows containing tagged entities, in more detail compared to main results in Figure \ref{fig:ne_loo}. G/M/N denote \emph{gold}, \emph{masked}, and \emph{names-only} retrieval targets. \emph{S} reports the paired \emph{swap} win rate of gold sentences over variants with replaced entities (chance $=50\%$).}
\label{tab:ne-control}
\end{table*}

\clearpage
\onecolumn
\section{Text Embedding Spaces}
\label{sec:text-spaces}



\begin{table}[H]
\centering

\begin{minipage}[c]{0.60\textwidth}
\centering
\setlength{\tabcolsep}{4.0pt}
\textbf{(a) Comparison of Text Spaces }\\[4pt]
\begin{tabular}{llrrr}
\toprule
\textbf{Speech encoder} & \textbf{Text target} & \textbf{Mono} & \textbf{Multi} & \textbf{LOO} \\
\midrule
\multirow{2}{*}{MMS-1B}
    & SONAR   & 26.13 & 30.46 & 15.41 \\
    & Harrier & 19.55 & 20.50 & 10.68 \\
\midrule
\multirow{2}{*}{w2v-BERT-2.0}
    & SONAR   & 35.52 & 47.37 & 32.57 \\
    & Harrier & 26.42 & 35.72 & 23.19 \\
\midrule
\multirow{2}{*}{XEUS}
    & SONAR   & 32.64 & 26.84 & 14.70 \\
    & Harrier & 27.94 & 19.57 & 11.07 \\
\midrule
\multirow{2}{*}{OmniASR-W2V-3B}
    & SONAR   & 18.90 & 23.89 & 19.92 \\
    & Harrier & 15.77 & 18.90 & 16.72 \\
\midrule
\multirow{2}{*}{WavLM Base+}
    & SONAR   & 12.50 & 10.11 & 7.93 \\
    & Harrier & 11.01 & 7.39 & 6.02 \\
\bottomrule
\end{tabular}
\end{minipage}%
\hfill
\begin{minipage}[c]{0.4\textwidth}
\centering
\textbf{(b) Correlations }\\[4pt]
\begin{tabular}{lrrr}
\toprule
 & Mono & Multi & LOO \\
\midrule
Pearson  & 0.99 & 0.99 & 0.98 \\
Spearman & 0.97 & 0.98 & 0.96 \\
\bottomrule
\end{tabular}
\end{minipage}

\caption{
Comparison of different text embedders, using SONAR and \texttt{harrier-oss-v1-0.6b} (similar in size to SONAR)
text targets on 39 non-English FLEURS languages, with the probes using the SONAR-selected layer for each speech encoder (layer selection is stable across both text spaces).
\textbf{(a):} Top-1 accuracy averaged across languages
for each speech encoder and probe training configuration.
\textbf{(b):} Pearson and Spearman correlations of probe
scores between text targets, with 195 observations each for
monolingual, multilingual, and leave-one-out (LOO) configurations.
Although SONAR achieves higher absolute accuracy, cross-lingual
and per-model probe behavior is strongly correlated. Thus, alternative text spaces mostly rescale the difficulty of the mapping, keeping cross-lingual and per-model behavior stable.
}
\label{tab:text_spaces}
\end{table}

\end{document}